\documentclass[runningheads]{llncs}
\usepackage[T1]{fontenc}
\usepackage{graphicx}
\usepackage{booktabs}
\usepackage[misc]{ifsym}
\newcommand{\corr}{(\Letter)}

\usepackage{mwe}
\usepackage{cite}
\usepackage{amsmath,amssymb,amsfonts}
\usepackage{algorithmic}
\usepackage{graphicx}
\usepackage{textcomp}
\usepackage{xcolor}
\usepackage{hyperref}
\usepackage{booktabs}
\usepackage{diagbox} 
\usepackage{multirow}
\begin{document}

\title{\textsc{FedRandom}: Sampling Consistent and Accurate\\ Contribution Values in Federated Learning.}

\titlerunning{FedRandom Contribution Values in Federated Learning}

\author{Arno Geimer \corr \and
Beltran Borja Fiz Pontiveros  \and
Radu State}

\authorrunning{A. Geimer et al.}

\institute{Interdisciplinary Centre for Security, Reliability and Trust  \\ University of Luxembourg \email{\{arno.geimer,beltran.fiz,radu.state\}@uni.lu}
}

\maketitle

\begin{abstract}
Federated Learning (FL) is a privacy-preserving decentralized approach for Machine Learning tasks. In industry deployments characterized by a limited number of entities possessing abundant data, the significance of a participant's role in shaping the global model becomes pivotal given that participation in a federation incurs costs, and participants may expect compensation for their involvement. Additionally, the contributions of participants serve as a crucial means to identify and address potential malicious actors and free-riders. However, fairly assessing individual contributions remains a significant hurdle. Recent works have demonstrated a considerable inherent instability in contribution estimations across aggregation strategies. While employing a different strategy may offer convergence benefits, this instability can have potentially harming effects on the willingness of participants in engaging in the federation. In this work, we introduce \textsc{FedRandom}, a novel mitigation technique to the contribution instability problem. Tackling the instability as a statistical estimation problem, \textsc{FedRandom} allows us to generate more samples than when using regular FL strategies. We show that these additional samples provide a more consistent and reliable evaluation of participant contributions. We demonstrate our approach using different data distributions across CIFAR-10, MNIST, CIFAR-100 and FMNIST and show that \textsc{FedRandom} reduces the overall distance to the ground truth by more than a third in half of all evaluated scenarios, and improves stability in more than 90\% of cases.

\end{abstract}

\section{Introduction}
\label{sec:intro}

Federated learning (FL) has emerged as a significant collaborative machine learning approach, gaining considerable attention for its ability to train models without requiring entities to share their potentially privacy-sensitive datasets.

Most of the work in FL traditionally focuses on optimising an evaluation metric (e.g global accuracy), convergence time and/or communication costs. The fairness of the participants' contribution to the global model is rarely evaluated or discussed. However, the equitable and accurate valuation of participant contributions is a critical challenge in FL, as deficiencies in this domain can severely undermine the integrity of the federated ecosystem. When participants perceive their contributions to be inadequately recognized or rewarded, their incentive to engage in the collaborative effort significantly diminishes \cite{bi2024understanding}. Consequently, contributors of high-quality data may exhibit reluctance or withdraw from the federation if they anticipate inequitable remuneration, and may even inadvertently foster malicious behaviours, such as rewarding participants who exhibit free-riding tendencies or submit falsified inputs \cite{lyu2020threats}.

Although the Shapley value has long been considered the standard for evaluating contributions \cite{ghorbani2019data}, recent work has shed light on an inherent instability in the consistent estimation of contributions by Shapley based methods\cite{geimer2025volatilityshapleybasedcontributionmetrics} and how this instability presents additional complexities to incentivize FL participants \cite{bi2024understanding}. In their work, the authors highlight the importance of the instability but provide no concrete method to address this instability.

The main contribution of this paper is the proposal of \textsc{FedRandom}, an aggregation technique to increase the sample size used in estimating the underlying true contribution of participants. It allows for the simulation of sample sets which significantly exceed the resulting sampling of the baseline strategy to approximate a centralized data valuation approach, and provides a mitigation strategy for the instability shown by Shapley-based contribution evaluation methods. The paper is structured as follows: We present a brief overview of the instability discovered in Shapley based contribution evaluation methods. Next, we present our method, \textsc{FedRandom}, its capabilities, as well as a comprehensive analysis of experimental results. Finally, we discuss the implications of contribution instabilities and the impact of \textsc{FedRandom} as a mitigation strategy.

\section{Background and Related Work}
While we do not dive into the specifics of Federated Learning, we detail a few keywords we use across this paper: An \textbf{aggregation strategy} denotes the technique used by the central server to obtain a new global model, examples are \textsc{FedAvg}\cite{mcmahan2017communication} or \textsc{FedProx}\cite{li2020federated}. The \textbf{epochs} $e$ are the amount of steps of gradient descent that participants compute locally. A \textbf{Dirichlet-based data split} is a quantity-based non-Independent and Identically Distributed (IID) split of the dataset across clients, with a concentration parameter $\alpha$ determining the non-iidness of the data (lower is less IID). 

\subsection{Contributions in FL}
Contribution methods in Federated Learning are used to determine weightings during aggregation, defend against malicious behavior, improve convergence, or incentivize and reward participation and efficient data usage. Overall, they can be divided into three main types:

\textbf{Self-reported} methods use information reported directly by the participants to the central server. This information can, for example, be used to rank clients\cite{mcmahan2017communication} with the goal of faster convergence, set the number of local steps taken \cite{wang2020tackling} or reduce energy consumption \cite{wiesner2024fedzero}. In a basic scenario using \textsc{FedAvg}, the server uses the sizes of the data set reported by the client to weigh model updates during the aggregation of the global model.

\textbf{Auction-based} contribution methods introduce an auction system, in which clients allow the central server to purchase their model updates at self-determined prices. This introduces a market mechanism to the contribution process, enabling the benefits to be not only dependent on the quality of the update, but also on the asking prices, which can lead to lower quality model updates being favored over more expensive high quality updates \cite{lu2023auction}\cite{pang2022incentive}. 

Finally, \textbf{Computation-based} methods involve a quality assessment, usually on the server side, of the participant's model updates. The most common contribution calculation method, borrowed from game theory, are Shapley values\cite{wang2019measure}\cite{wang2020principled}. Their purpose is to assess the contribution of each participant by comparing the outcomes that include the participant with those that exclude them. Although this method facilitates an equitable distribution of the overall contribution among participants, taking into account their individual influence on the final model, it is computationally expensive due to the combinatorial nature of the approach. 

In this paper, we focus on the computation-based Multi-Round Reconstruction Shapley values\cite{song2019profit}, where the importance of model updates is determined each round. For comparisons, we normalize contribution values.

\subsection{Contribution instability in FL}

Although different contribution methods in Federated Learning have been widely studied, \cite{geimer2025volatilityshapleybasedcontributionmetrics} observes a contribution instability across different aggregation strategies. Namely, the paper gives an evaluation of participant contributions using gradient-based model reconstruction techniques with Shapley values. The paper notes significantly unstable reward allocations for clients across different FL strategies. This observation has important implications for federations. Rather than agreeing on a common aggregation method, participants may want to optimize their personal reward, leading to disagreement when choosing a strategy and non-participation due to perceived unfair contributions. Furthermore, the authors make a crucial observation concerning the performance of each strategy. There is no clear winner in approximating a baseline, no method consistently outperforms the others.

\label{sec:methodology}
\section{Methodology}

Since Shapley values are a purely computational means of estimating contributions, results of every aggregation strategy are inherently valid. However, as a consequence of the high variation in-between strategies and lack of an optimal strategy, there is no clear argument for the best contribution calculation method. A central server's choice of evaluation method would likely lead the least rewarded client to criticize the fairness of the selection. 

A common way to extrapolate information from a set of independent noisy data is to use statistical methods. We assume contributions of different strategies to be noisy samples following an underlying real contribution distribution. Treating this as a statistical estimate of the underlying true contribution, we use the arithmetic mean of the samples as an estimator for the underlying contribution. Coincidentally, the principle of using the sample mean as an estimator of a representative quantity is the foundation of Federated Learning, which uses an aggregation of model updates as the estimator for a global model. In this light, we propose two methods for sampling contributions, discuss their pros and cons, and showcase the setup of our experiments.

\subsection{MSM sampling}

First, we collect different contribution samples using a group of aggregation strategies, namely \textsc{FedAvg}\cite{mcmahan2017communication}, \textsc{FedAvgM}\cite{hsu1909measuring}, \textsc{FedAdagrad}, \textsc{FedAdam} and \textsc{FedYogi}\cite{reddi2020adaptive}, \textsc{FedMedian} and \textsc{FedTrimmedAvg}\cite{yin2018byzantine} and \textsc{Krum}\cite{blanchard2017machine}. The overall contribution is determined as the mean of the sample. We call this means of determining contributions the mean strategy method, or MSM. Given that the mean of contribution percentages sums up to one and is thus a percentage representation of contributions itself, assessing contributions this way reduces friction between clients, since a number of valid contributions are used equally to determine final reward payments. 

Although MSM is a straightforward way of dealing with the instability, the method has a substantial shortcoming, in that it requires the implementation and execution of different aggregation techniques, which leads to significant preparatory efforts. As each generated sample is based on a new aggregation technique, it is not scalable. Consequently, it is not possible to achieve high confidence levels, as the method cannot produce a sufficient number of samples, since, even when ignoring implementation overhead, there is only a limited supply of different aggregation strategies in Federated Learning.

\subsection{\textsc{FedRandom}}\label{sec:FedRandom}
Extending the idea of sampling multiple contribution measures, we introduce \textsc{FedRandom} to generate samples from an ensemble of aggregation techniques by using a randomized process. Based on already-established aggregation techniques, it is able to simulate more extensive sample sets.

More specifically, we use aggregation strategies $S = $ (\textsc{FedAvg}, \textsc{FedAvgM}, \textsc{FedAdagrad}, \textsc{FedAdam}, \textsc{FedYogi}), the best performing methods in \cite{geimer2025volatilityshapleybasedcontributionmetrics}. Although the overall process mirrors a basic federation with \textsc{FedAvg}, during the aggregation step, \textsc{FedRandom} uniformly samples one of the aggregation techniques of $S$ and uses it as the server-side aggregation function. The federation continues as usual: The server distributes the newly aggregated model to the clients who train it on their local data and send updates to the server. Any strategy-specific quantities related to the chosen aggregation strategies, such as momentum, are determined each round with minimal computational load.

Due to its randomness, \textsc{FedRandom} allows the simulation of a vast amount of strategies, potentially producing more contribution values than MSM, while being implemented as a single aggregation technique. Although the amount of strategies used to determine a mean percentage of contributions with MSM is limited, sampling with \textsc{FedRandom} is almost boundless. In fact, it enables the simulation of up to $s^r$ different Federated Learning processes, where $r$ is the number of communication rounds and $s = |S|$.
We will show that the use of contribution means with data produced using \textsc{FedRandom} addresses the observed instability of individual aggregation strategies in contribution calculation. It will enable better and more consistent contribution estimates compared to MSM.

\subsection{Cost of sampling with \textsc{FedRandom}}

Statistically improving estimators by expanding sample sets always comes with a drawback: The cost of collecting new samples. The trade-off between training data size and the accuracy of a final model is one of the most important challenges in building any statistical model. Indeed, one of the arguments for implementing Federated Learning solutions is the poorer performance of ML models trained on too small datasets. In light of this, the additional computation and communication cost caused by collecting more extensive sample sets using \textsc{FedRandom} follows a similar argument: Federations have to assess whether the improved contribution estimation is worth the additional cost. However, reward sharing is a more likely scenario for cross-silo federations, which include few participants with strong hardware capabilities. Thus, we argue that the additional computational cost incurred by sampling is set off by the improved contribution stability and trust in the federation.
\subsection{Experimental setup}

We replicated the experimental setup from \cite{geimer2025volatilityshapleybasedcontributionmetrics}. We use the same computer vision datasets ( \textsc{Cifar-10}, \textsc{Cifar-100}\cite{krizhevsky2009learning}, \textsc{Mnist}\cite{deng2012mnist} and Fashion-\textsc{Mnist}\cite{xiao2017fashion}) with the same amount of clients and training steps. We use the same CNN architecture, consisting of two 5x5 convolution layers, 2 dense layers with ReLu activation and 16x5x5x120 and 120x84 units, respectively, and a dense layer with 84x10 units. We use the same concentration parameters for Dirichlet-based splits, $\alpha = (1, 10, 100)$. Our experiments include 9 different seeds, leading to a total of 324 = (3 epochs $\times$ 4 datasets $\times$ 3 $\alpha$-values $\times$ 9 seeds) different Federated Learning scenarios. Each scenario contains a federation using the 8 aggregation strategies present in MSM, as well as 30 \textsc{FedRandom} runs. This gives a total of over 12.000 federations, all simulated on V100 16GB and 32GB GPUs.

\section{Results}\label{sec:results}

This section includes a brief analysis of the convergence of \textsc{FedRandom} as well as an extensive study of the validity of examples generated using MSM and \textsc{FedRandom}, with the goal of demonstrating the superior quality of using \textsc{FedRandom} to sample contributions.

\subsection{\textsc{FedRandom} as an aggregation technique}

\begin{figure}[h]
\begin{center}
    
   \includegraphics[scale = .5]{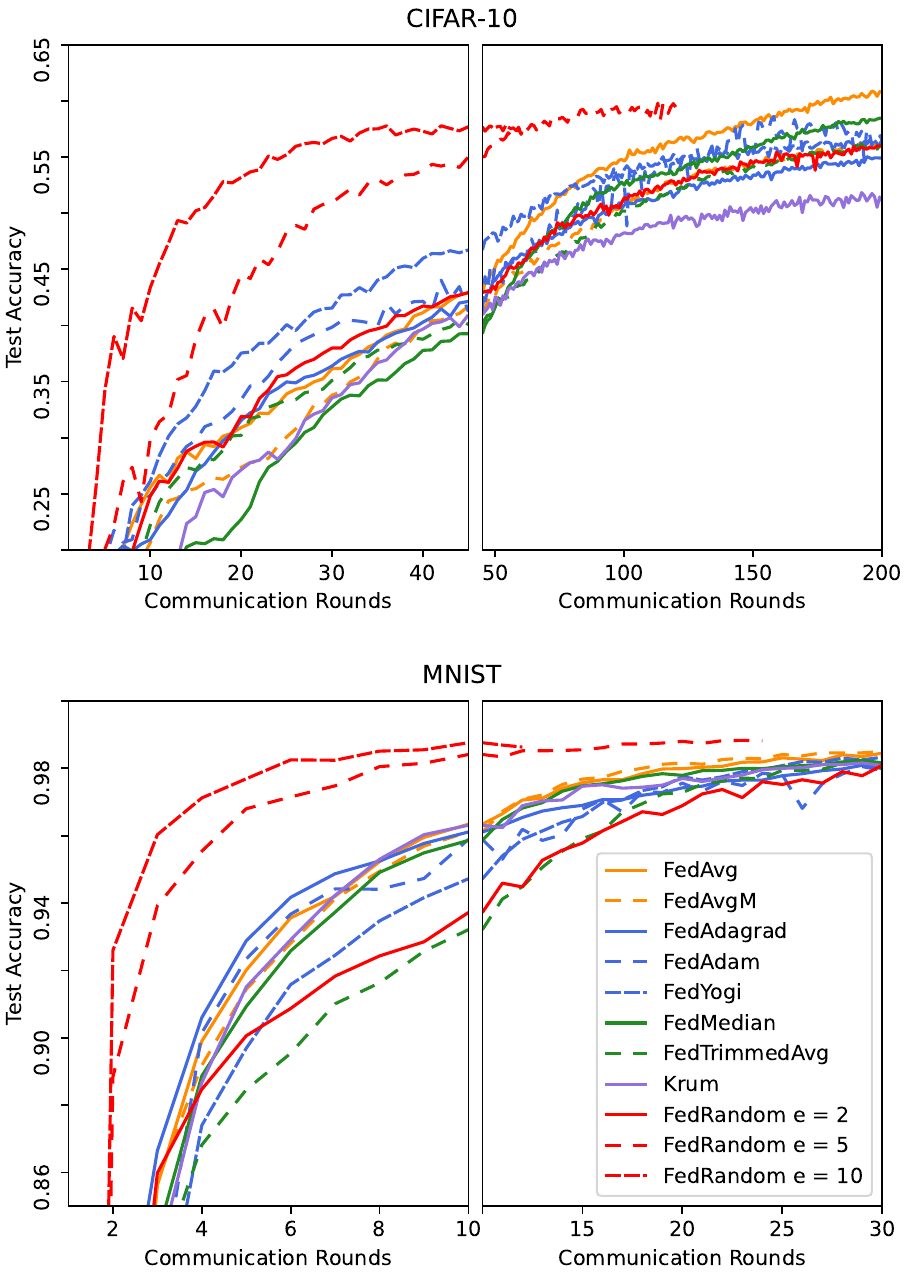}
   \caption{Test set accuracy on CIFAR-10 and MNIST for $e =$ 2. \textsc{FedRandom} convergence speed lies between that of baseline aggregation strategies. \textsc{FedRandom} $e = $( 5, 10) is included to visualize convergence under different parameters.}
   \label{fig:FRacc}
    \end{center}
\end{figure}

All aggregation strategies which make up \textsc{FedRandom} share the same objective function, therefore it can be considered a version of \textsc{FedAvg} with momentum or an adaptive learning rate applied at random rounds. Figure \ref{fig:FRacc} shows the convergence speed of \textsc{FedRandom} compared to the baseline strategies.

It is important to note that while \textsc{FedRandom} serves as a valid aggregation strategy, its performance in terms of model convergence, as illustrated in Figure \ref{fig:FRacc}, is not demonstrably superior to established methods. Consequently, its primary utility lies in the domain of contribution estimation. This positions \textsc{FedRandom} as a specialized tool where its computational cost must be weighed against the gains in the stability and perceived accuracy of contribution values. If a different, potentially higher-performing, aggregation strategy is used for the primary federated learning task, the use of \textsc{FedRandom} for contribution analysis represents an additional computational overhead that must be justified by the value derived from more consistent and reliable contribution assessments.

\subsection{\textsc{FedRandom} for variance and bias minimization}

\begin{figure}[h]
    \begin{center}
   \includegraphics[scale = .5]{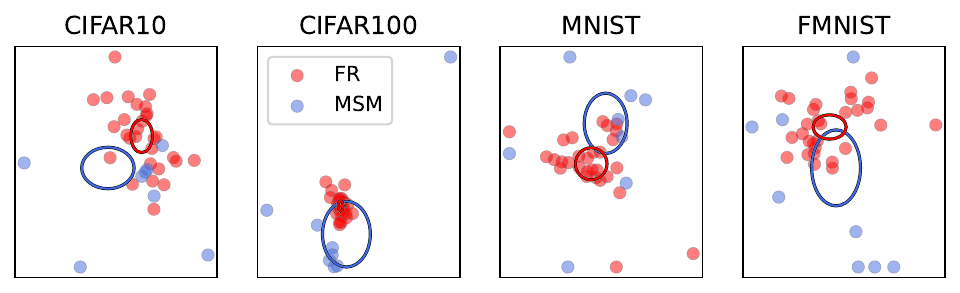}
   \caption{2D visualization of the variance in contributions sampled with regular aggregation techniques (\textcolor{blue}{blue}), as well as \textsc{FedRandom} (\textcolor{red}{red}). Ellipses show standard deviations. To note that the centers of the ellipses, as the average of a bundle of examples, are the MSM and \textsc{FedRandom} sampling result.}
   \label{fig:scatt_var}
    \end{center}
\end{figure}

Starting with variance minimization, Figure \ref{fig:scatt_var} demonstrates how \textsc{FedRandom} enables more consistent contribution allocations compared to MSM: First, it shows \textsc{FedRandom}'s superior capability to generate a greater amount of contribution samples, which MSM is unable to do without implementing new aggregation strategies; second, \textsc{FedRandom} contributions also have substantially smaller standard deviation compared to MSM, whose allocations appear far more spread out.

\begin{table*}[!htbp]
\hspace*{-0.6cm}
    \begin{tabular}{c c c|c|c| |c|c|c| |c|c|c| |c|c|c}
        \cline{3-14}
         \multicolumn{2}{c}{} & \multicolumn{3}{c||}{\textsc{Cifar-10}} & \multicolumn{3}{c||}{\textsc{Cifar-100}} & \multicolumn{3}{c||}{\textsc{Mnist}} & \multicolumn{3}{c}{\textsc{FMnist}} \\
        \cline{2-14} 
        & \multicolumn{1}{|c|}{\diagbox[height=1.5\line]{$e$}{$\alpha$}} & 1 & 10 & 100  & 1 & 10 & 100 & 1 & 10 & 100 & 1 & 10 & 100 \\
        \hline
        \hline
        \multicolumn{1}{c|}{MSM} & \multirow{2}{*}{2} & \multicolumn{1}{|c|}{0.059} & 0.056 &0.057 &0.069 &0.064 &0.066 &0.052 &0.06 &0.069 &0.046 &0.058 &0.069 \\
        \multicolumn{1}{c|}{FR}  &  & \multicolumn{1}{|c|}{\textbf{0.006}} &\textbf{0.003} &\textbf{0.005} &0.003 &\textbf{0.003} &\textbf{0.005} &\textbf{0.01} &\textbf{0.014} &\textbf{0.014} &\textbf{0.007} &\textbf{0.013} &\textbf{0.016} \\ 
        \hline
        \multicolumn{1}{c|}{MSM} & \multirow{2}{*}{5} & \multicolumn{1}{|c|}{0.057} &0.051 &0.067 &0.054 &0.055 &0.059 &0.056 &0.061 &0.072 &0.045 &0.062 &0.072 \\
        \multicolumn{1}{c|}{FR}  & & \multicolumn{1}{|c|}{\textbf{0.006}} &0.004 &0.007 &\textbf{0.002} &\textbf{0.003} &\textbf{0.005} &0.012 &\textbf{0.014} &0.017 &0.01 &0.014 &0.017 \\
        \hline
        \multicolumn{1}{c|}{MSM} & \multirow{2}{*}{10} & \multicolumn{1}{|c|}{0.05} &0.042 &0.054 &0.05 &0.05 &0.059 &0.057 &0.067 &0.074 &0.047 &0.061 &0.068 \\
        \multicolumn{1}{c|}{FR}  & & \multicolumn{1}{|c|}{0.007} &0.006 &0.011 &0.003 &\textbf{0.003} &\textbf{0.005} &0.012 &0.015 &0.016 &0.011 &0.015 &0.017 \\
        \hline
        \multicolumn{14}{c}{} \\

       \end{tabular}
       \caption{Average standard deviation in contribution samples generated by MSM and by \textsc{FedRandom} (FR), lower is better. \textbf{Bold} entries are best in column.}
       \label{table:var}
\end{table*}

While we have shown that sampling with \textsc{FedRandom} greatly decreases variance when comparing different aggregation strategies, this does not necessarily mean that \textsc{FedRandom} gives a realistic approximation of clients' contributions. In accordance with previous research \cite{yu2020sustainable}\cite{guo2024fair}\cite{shyn2022empirical}, we use size-based contributions determined by the sizes of different clients' datasets as a baseline ground truth.
We compare the contribution allocations generated using MSM, the mean of the baseline strategies, and those generated using the mean of several \textsc{FedRandom} runs, to this ground truth.

\begin{figure}[h]
    \begin{center}
   \includegraphics[scale = .5]{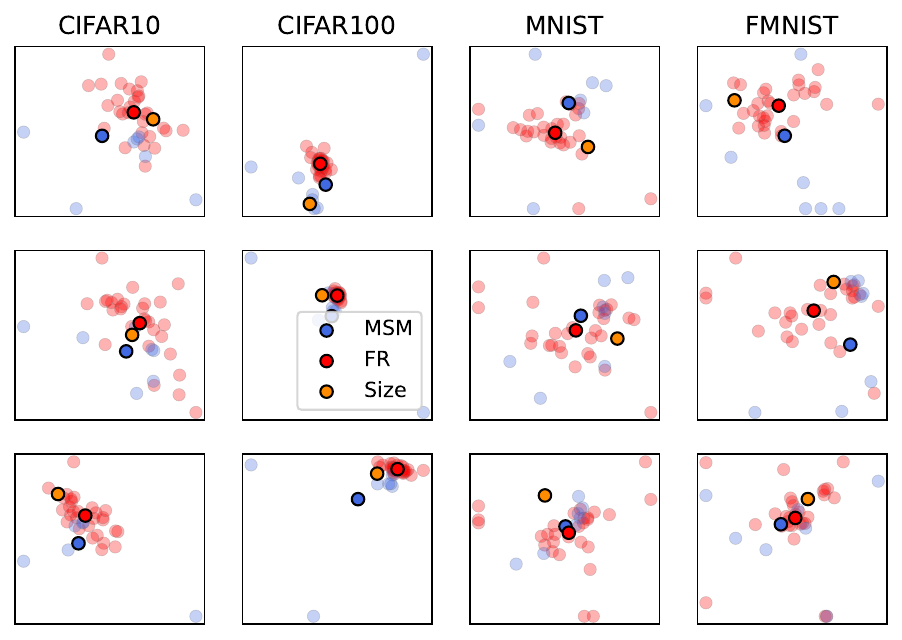}
   \caption{2D visualization of average contributions sampled using MSM (\textcolor{blue}{blue}), as well as \textsc{FedRandom} (\textcolor{red}{red}), compared to the size-based baseline (orange).}
   \label{fig:scatt_dist}
   \end{center}
\end{figure}

\begin{table*}[!htbp]
\hspace*{-0.6cm}
    \begin{tabular}{cc c|c|c| |c|c|c| |c|c|c| |c|c|c}
        \cline{3-14}
        &  & \multicolumn{3}{c||}{\textsc{Cifar-10}} & \multicolumn{3}{c||}{\textsc{Cifar-100}} & \multicolumn{3}{c||}{\textsc{Mnist}} & \multicolumn{3}{c}{\textsc{FMnist}} \\
        \cline{2-14}
        \multicolumn{1}{c|}{}        & \multicolumn{1}{c|}{\diagbox[height=1.5\line]{$e$}{$\alpha$}} & 1 & 10 & 100  & 1 & 10 & 100 & 1 & 10 & 100 & 1 & 10 & 100 \\
        \hline
        \hline
        \multicolumn{1}{c|}{MSM}       & \multirow{2}{*}{2} & \multicolumn{1}{|c|}{0.245} & 0.299 & 0.349 & 0.206 & 0.359 & 0.268 & 0.147 & 0.165 & 0.176 & 0.156 & 0.178 & 0.256 \\
        \multicolumn{1}{c|}{FR}       &  & \multicolumn{1}{|c|}{0.214} & 0.26 & \textbf{0.342} & 0.148 & 0.346 & 0.255 & 0.099 & \textbf{0.12} & 0.111 & \textbf{0.143} & \textbf{0.128} & 0.221\\
        \hline
        \multicolumn{1}{c|}{MSM} & \multirow{2}{*}{5} & \multicolumn{1}{|c|}{0.223} & \textbf{0.245} & 0.37 & 0.198 & 0.344 & 0.251 & 0.133 & 0.201 & 0.137 & 0.16 & 0.19 & 0.237 \\
        \multicolumn{1}{c|}{FR}     &  & \multicolumn{1}{|c|}{0.214} & 0.253 & \textbf{0.342} & 0.146 & 0.343 & 0.242 & 0.087 & 0.126 & 0.118 & 0.153 & 0.142 & \textbf{0.213} \\
        \hline
        \multicolumn{1}{c|}{MSM}       & \multirow{2}{*}{10} & \multicolumn{1}{|c|}{\textbf{0.212}} & 0.269 & 0.344 & 0.181 & \textbf{0.321} & 0.252 & 0.112 & 0.175 & 0.15 & 0.177 & 0.161 & 0.22 \\
        \multicolumn{1}{c|}{FR}       &  & \multicolumn{1}{|c|}{\textbf{0.212}} & 0.247 & 0.346 & \textbf{0.142} & 0.349 & \textbf{0.241} & \textbf{0.075} & 0.13 & \textbf{0.108} & \textbf{0.143} & 0.15 & 0.218 \\
        \hline
        
        \multicolumn{14}{c}{} \\
        \hline
        \multicolumn{1}{c|}{MSM}       & \multirow{2}{*}{2} & \multicolumn{1}{|c|}{0.191} & 0.237 & 0.275 & 0.159 & 0.282 & 0.214 & 0.104 & 0.123 & 0.128 & 0.118 & 0.133 & 0.193 \\ 
        \multicolumn{1}{c|}{FR}       &  & \multicolumn{1}{|c|}{\textbf{0.163}} & 0.203 & \textbf{0.267} & 0.117 & 0.273 & 0.199 & 0.068 & \textbf{0.084} & \textbf{0.08} & 0.109 & 0.096 & 0.159 \\
        \hline
        \multicolumn{1}{c|}{MSM} & \multirow{2}{*}{5} & \multicolumn{1}{|c|}{0.174} & 0.193 & 0.289 & 0.155 & 0.271 & 0.198 & 0.098 & 0.132 & 0.098 & 0.119 & 0.147 & 0.178 \\
        \multicolumn{1}{c|}{FR}     & & \multicolumn{1}{|c|}{0.164} & 0.198 & 0.269 & 0.114 & 0.271 & 0.19 & 0.066 & 0.091 & 0.081 & 0.112 & \textbf{0.106} & \textbf{0.154} \\
        \hline
        \multicolumn{1}{c|}{MSM}       & \multirow{2}{*}{10} & \multicolumn{1}{|c|}{0.165} & 0.215 & 0.265 & 0.144 & \textbf{0.254} & 0.197 & 0.084 & 0.12 & 0.111 & 0.132 & 0.126 & 0.16 \\
        \multicolumn{1}{c|}{FR}       &  & \multicolumn{1}{|c|}{0.164} & \textbf{0.19} & 0.272 & \textbf{0.112} & 0.276 & \textbf{0.19} & \textbf{0.055} & 0.093 & 0.081 & \textbf{0.104} & \textbf{0.106} & 0.158 \\
        \hline
        \multicolumn{14}{c}{} \\

       \end{tabular}
       \caption{Average $\mathcal{L}_2$ (top) and $\mathcal{L}_\infty$ (bottom) distance between the baseline and contributions generated by MSM and \textsc{FedRandom} (FR), lower is better. \textbf{Bold} entries are best in column by metric.}
       \label{table:means}
\end{table*}

This is visualized in Figure \ref{fig:scatt_dist}: When plotting mean per-client contributions calculated using both methods against a size-based percentage, we observe that \textsc{FedRandom} tends to outperform MSM in estimating contributions. 

\subsection{\textbf{\textsc{FedRandom}} in contribution evaluation}
To assess the performance of both methods, we compare their effectiveness in reducing both variance, i.e. deviation between samples, as well as bias, the closeness to the ground truth. In contrast to the instability observed earlier, a generalized and stable contribution evaluation is desirable in Federated Learning. We employ \textsc{FedRandom} as a sample augmentation technique to address the shortcomings of MSM.

This observation is supported by Table \ref{table:var}, which comprises the average standard deviation $\sigma$ of MSM and of \textsc{FedRandom} contributions. We observe that \textsc{FedRandom} allows the sampling of drastically less variant examples. Consequently, \textsc{FedRandom} results in a more robust and consistent contribution allocation, leading to more confidence in individual contributions. We are able to decrease the standard deviation by at least a factor of 4, with certain examples showing a deviation improvement by a factor of 40.

Indeed, this observation is supported by our extensive study, including a multitude of data splits as well as epoch values, in Table \ref{table:means}. We showcase that \textsc{FedRandom} does indeed substantially outperform a baseline MSM contribution allocation in nearly all settings. To assess overall closeness to the ground truth, we compute the average $\mathcal{L}_2$-distance to the baseline, while we employ the $\mathcal{L}_\infty$-distance to assess the worst per-client divergence.
While the table makes it obvious that \textsc{FedRandom} significantly outperforms MSM, there are several observations we discuss. First, our method surpasses MSM in almost all settings, only being outmatched in a few data splits on \textsc{Cifar}-10 for some epoch values.
Additionally, it allows us to reduce the overall distance to the ground truth by more than a third in half of all scenarios, with some outperforming MSM by more than 35\%. In total, we beat MSM in 33 out of 36 (92\%) of cases. In all of these cases, we beat MSM in worst per-client performance ($\mathcal{L}_\infty$) as well. These substantial improvements are a strong argument towards the adaptation of \textsc{FedRandom} sampling when establishing a cross-silo federation with contribution calculation.

The validity of the experimental results are clear when analysing the statistics over all 324 scenarios. \textsc{FedRandom} has lower variance than MSM in 303 (94\%) cases, lower $\mathcal{L}_2$-distance to the baseline in 275 (85\%) of cases, and lower $\mathcal{L}_\infty$-distance in 267 (82\%) of cases. P-values show that \textsc{FedRandom} outperforms MSM with probability almost 1.

In summary, we have shown that not only are \textsc{FedRandom} samples less dispersed, they also allow us to better estimate ground truth contribution allocations. The conclusion contains an in-depth discussion on the implications of these findings.

\section{Conclusion}
Given the previously observed instability in contributions across strategies, our proposed method, \textsc{FedRandom}, allows us to significantly reduce variance and bias in determining contributions compared to baseline MSM. Sampling with \textsc{FedRandom} allows central servers to estimate contributions before running a federation. Not only are these contribution values more consistent than baseline sampling with MSM, they are also closer to the ground truth. Besides, \textsc{FedRandom} does not require a vast amount of different aggregation techniques to be implemented.
In a deployed cross-silo federation, employing our method reduces friction between participants: As opposed to before, where a specific aggregation strategy might have been more beneficial to a client, statistically relevant contribution values can be sampled ahead. This leaves the option to choose the strategy which is most beneficial to the federation as a whole, eliminating inter-client concurrence. Participants getting rewarded more fairly when employing \textsc{FedRandom} greatly improves trust in the remuneration process.

Naturally, any sampling introduces additional computational effort to a federation. Be it with \textsc{FedRandom} or with MSM, this effort scales linearly with the amount of generated samples. Additionally, due to their nature, the computational effort of Shapley values scales exponentially with the number of clients, rendering simulations in high-client environments highly demanding.

In future works, the effect of different aggregation strategies in \textsc{FedRandom}, as well as the importance of the strategy types should be studied. Involved strategies do not necessarily need to share the same objective function, and these cases should be tested. The use of alternative contribution methods with \textsc{FedRandom} should be considered, since a suitable validation dataset might not always be available.
Furthermore, the possible benefits of randomly mixing strategies, as proposed by \textsc{FedRandom}, should be investigated in a more general context. Randomization during aggregation might be of particular interest in security, making aggregation strategy fingerprinting harder.


\textsc{FedRandom}'s approach to stable contribution assessment will prove increasingly valuable as FL tackles more sophisticated challenges ahead. For instance, in emerging FL tasks such as the federated fine-tuning of Large Language Models (LLMs) or collaborative generative AI, fairly evaluating each participant's contribution to the model's capabilities (e.g., safety profile, specific knowledge domains, or creative output) will be paramount. The inherent complexity of assessing such diverse contributions may exacerbate existing instability issues. This underscores the critical importance of methodologies like \textsc{FedRandom} to ensure fairness and maintain trust as FL expands into these domains.

\bibliographystyle{splncs04}
\bibliography{main}
\end{document}